\documentclass[10pt,journal,cspaper,compsoc]{IEEEtran}
\ifCLASSOPTIONcompsoc
\else
\fi
\ifCLASSINFOpdf
  \usepackage[pdftex]{graphicx}
\else
\fi
\usepackage{url}

\usepackage{tshlee}

\begin{document}
%
\title{A Framework for Symmetric Part Detection in Cluttered Scenes}

\author{Tom~Lee,
        Sanja~Fidler,
        Alex~Levinshtein,
        Cristian~Sminchisescu,
        and~Sven~Dickinson
\IEEEcompsocitemizethanks{\IEEEcompsocthanksitem T. Lee, S. Fidler, and S. Dickinson are with the Department of Computer Science, University of Toronto.\protect\\
Email: $\tt{\{tshlee,fidler,sven\}@cs.toronto.edu}$
\IEEEcompsocthanksitem A. Levinshtein is with Epson.
\IEEEcompsocthanksitem C. Sminchisescu is with Lund University.}
\thanks{}}

\IEEEcompsoctitleabstractindextext{%
\begin{abstract}
The role of symmetry in computer vision has waxed and waned in importance during the evolution of the field from its earliest days.  At first figuring prominently in support of bottom-up indexing, it fell out of favor as shape gave way to appearance and recognition gave way to detection.  With a strong prior in the form of a target object, the role of the weaker priors offered by perceptual grouping was greatly diminished. However, as the field returns to the problem of recognition from a large database, the bottom-up recovery of the parts that make up the objects in a cluttered scene is critical for their recognition.  The medial axis community has long exploited the ubiquitous regularity of symmetry as a basis for the decomposition of a closed contour into medial parts.  However, today's recognition systems are faced with cluttered scenes, and the assumption that a closed contour exists, \ie that figure-ground segmentation has been solved, renders much of the medial axis community's work inapplicable.  In this article, we review a computational framework, previously reported in \cite{Lee13Detecting, Levinshtein09Multiscale, Levinshtein13Multiscale}, that bridges the representation power of the medial axis and the need to recover and group an object's parts in a cluttered scene.  Our framework is rooted in the idea that a maximally inscribed disc, the building block of a medial axis, can be modeled as a compact superpixel in the image.  We evaluate the method on images of cluttered scenes.
\end{abstract}

\begin{keywords}
symmetry; medial axis; perceptual grouping; object recognition
\end{keywords}}

\maketitle

\IEEEdisplaynotcompsoctitleabstractindextext

%
\IEEEpeerreviewmaketitle

\section{Introduction}

\IEEEPARstart{S}{ymmetry} is a long-standing, interdisciplinary form that spans across the arts and sciences, covering fields as disparate as mathematics, biology, architecture, and music \cite{Leyton92Symmetry}.  The roles played by symmetry are equally diverse, and can involve being an abstract object of analysis, a balancing structure in nature, or an attractor of visual attention.  The common thread in all of the above is that symmetry is ubiquitously present in both natural objects and artificial objects.  It is no accident that we constantly encounter symmetry through our eyesight, and in fact, Gestalt psychologists \cite{Wertheimer38Laws} of the previous century proposed that symmetry is a physical regularity in our world that has been exploited by the human visual system to yield a powerful perceptual grouping mechanism.  Experiments show evidence that we respond to symmetry before being consciously aware of it \cite{Tyler02Human}.

The scope of this article lies within the domain of \emph{computer vision}, a comparatively young field that has adopted symmetry since its infancy.  Inspired by a computational understanding of human vision, perceptual grouping played a prominent role in support of early object recognition systems, which typically took an input image and a set of shape models, and identified which of the models was visible in the image.  Mid-level shape priors were crucial in grouping causally related shape features into discriminative shape indices that were used to prune the set down to a few promising candidates that might account for a query.  Of these shape priors, one of the most powerful is a configuration of parts, in which a set of related parts belonging to the same object is recovered without any prior knowledge of scene content.

The use of symmetry to recover generic parts from an image can be traced back to the earliest days of computer vision, and includes the medial axis transform (MAT) of Blum (1967) \cite{Blum67Transformation}, generalized cylinders of Binford (1971) \cite{Binford71Visual}, superquadrics of Pentland (1986) \cite{Pentland86Perceptual}, and geons of Biederman (1985) \cite{Biederman85Human}, to name just a few examples.  Central to a large body of approaches based on \emph{medial symmetry} is the MAT, which decomposes a closed 2D shape into a set of connected medial branches corresponding to a configuration of parts, providing a powerful parts-based decomposition of the shape suitable for shape matching, \eg Siddiqi \etal. (1999) \cite{Siddiqi99Shock} and Sebastian \etal. (2004) \cite{Sebastian04Recognition}.  For a definitive survey on medial symmetry, see Siddiqi \etal. (2008) \cite{Siddiqi08Medial}.

In more recent years, the field of computer vision has shifted in focus toward the object detection problem, in which the input image is searched for a specific target object.  One reason for this lies in the development of machine learning algorithms that can leverage large amounts of training data to produce robust classification results.  This led to rapid progress in the development of object detection systems, enabling them to handle increasing levels of background noise, occlusion, and variability in input images \cite{Girshick14Rich}.  This development established the standard practice of working with input domains of real images of cluttered scenes, significantly increasing the applicability of object recognition systems to real problems. 

\newcommand{\PartRep}{PartRep}
\newcommand{\Multiscale}{Multiscale}

\begin{figure}[t]
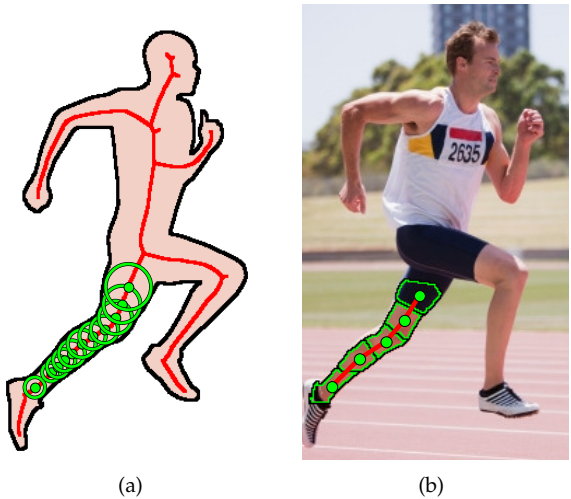

\newcommand{\GetIt}[1]{\includegraphics[height=6cm, trim=2mm 2 2mm 2mm, clip]{includegraphics/\PartRep-#1.pdf}}
\centering
\subfigure[]{\GetIt{00}}
\subfigure[]{\GetIt{02}}
\caption{Our representation of symmetric parts: In (a), the shape of the runner's body is transformed into its medial axis (red), a skeleton-like structure that decomposes the shape into branch-like segments, \eg the leg.  The leg's shape is swept out by a sequence of discs (green) lying along the medial axis.  In (b), the shape of the same leg is composed from superpixels that correspond to the sequence of discs.  The scope of this article's framework is limited to detecting symmetric parts corresponding to individual branches.}
\label{fig:\PartRep}
\end{figure}

A parallel advance in perceptual grouping, however, did not occur for a simple reason: With the target object already known, indexing is not required to select it, and perceptual grouping is not required to construct a discriminative shape index.  As a result, perceptual grouping activity at major conferences has diminished along with the supporting role of symmetry \cite{Dickinson09Evolution,Dickinson13Role}.  However, there are clear signs that the object recognition community is moving from appearance back to shape, and from object detection back to multiclass object categorization.  Shape-based perceptual grouping will play a critical role in facilitating this transition.

In attempting to bring back medial symmetry in support of perceptual grouping, we observe that the subcommunity's efforts have not kept pace with mainstream object recognition.  Specifically, medial symmetry approaches typically assume that the input image is a foreground object free from occluding and background objects and, accordingly, lack the ability to segment foreground from background, an ingredient crucial for tackling contemporary datasets.  It is clear that the MAT cannot be reintroduced without combining it with an approach for figure-ground segmentation.  In this article, we review current work along this trajectory as represented by Lee \etal. (2013) \cite{Lee13Detecting} and earlier work by Levinshtein \etal. (2009) \cite{Levinshtein09Multiscale,Levinshtein13Multiscale}.

In the context of symmetric part detection, \cite{Lee13Detecting} introduced an approach that leveraged earlier work \cite{Levinshtein09Multiscale} to build a MAT-based superpixel grouping method.  Since the proposed representation is central to our approach, we proceed with a brief overview of the latter.  A bottom-up method was introduced which first detected symmetric parts, then grouped them non-accidentally to form a discriminative shape index.  By establishing a correspondence between superpixels and maximally inscribed discs, the method formulated a superpixel grouping problem that exploited symmetry as a grouping cue.  The method thus recovered symmetric parts by grouping superpixels that represented discs of the same part.  An evaluation was presented to show a significant improvement over other symmetry-based approaches.

Subsequently, \cite{Lee13Detecting} furthered the development of the above ideas on two complementary fronts.  First, the medial representation was used to derive a sequence optimization problem for grouping, whose solution was shown to bring significant improvements in results.  The approach uses a grouping algorithm that is principled and more effective than in \cite{Levinshtein09Multiscale}.  Second, symmetry was captured more accurately by increasing the number of model parameters.  While a limited number of parameters previously captured scale and orientation, the method's invariance was improved by additionally capturing bending and tapering.  The resulting affinity function was also shown to support an improvement.

This article takes a high-level view of the work in reintroducing the MAT with figure-ground segmentation capability, enabling us to draw insights from a higher vantage point.  We first develop the necessary background to trace the development from its origins in the MAT, through \cite{Levinshtein09Multiscale}, and finally to \cite{Lee13Detecting}.  In doing so, we establish a framework that makes clear the connections among previous work.  For example, it follows from our exposition that \cite{Levinshtein09Multiscale} is an alternative instance of our framework.  More generally, our unified framework benefits from the rich structure of the MAT while directly tackling the challenge of segmenting out background noise in a cluttered scene.  Our model is discriminatively trained and stands out from typical perceptual grouping methods that use predefined grouping rules.  Using experimental image data, we present both qualitative results and a quantitative metric evaluation to support the development of the components of our approach.

\section{Related work}

Symmetry is one of several important Gestalt cues that contribute to perceptual grouping.  Symmetry plays neither an exclusive nor an isolated role in the presence of other cues.  Contour closure, for example, is another mid-level cue whose role will increase as the community relies more on bottom-up segmentation in the absence of a strong object prior, \eg \cite{Levinshtein12Optimal}. Symmetry may also be effectively combined with other mid-level cues, \eg \cite{Ren05Cue, Lee14Multicue}.  For brevity, we restrict our survey of related work to symmetry detection.

The MAT, along with its many descendant representations such as the shock graph \cite{Sebastian04Recognition,Siddiqi99Shock,Pelillo99Matching,Demirci09Skeletal} and bone graph \cite{Macrini11Object,Macrini11Bone}, provides an elegant decomposition of an object's shape into symmetric parts; however, it made the unrealistic assumption that the shape was segmented, and is thus not directly suitable for today's image domains.  For symmetry approaches in the cluttered image domain, we first consider the \emph{filter-based} approach, which first attempts to detect local symmetries, in the form of parts, and then finds non-accidental groupings of the detected parts to form indexing structures.  Example approaches in this domain include the multiscale peak paths of Crowley \& Parker (1984) \cite{Crowley84Representation}, the multiscale blobs of
Shokoufandeh \etal. (1999) \cite{Shokoufandeh99Viewbased},
the ridge detectors of Mikolajczyk \& Schmid (2002) \cite{Mikolajczyk02Affine}, and the multiscale blobs and ridges of Lindeberg \& Bretzner (2003) \cite{Lindeberg03Realtime}, and Shokoufandeh \etal. (2006) \cite{Shokoufandeh06Representation}.  Unfortunately, these filter-based approaches yield many false positive and false negative symmetric part detections, and the lack of explicit part boundary extraction makes part attachment detection unreliable.

A more powerful filter-based approach was recently proposed by Tsogkas \& Kokkinos (2012) \cite{Tsogkas12Learningbased}, in which integral images are applied to an edge map to efficiently compute discriminating features, including a novel spectral symmetry feature, at each pixel at each of multiple scales.  Multiple instance learning is used to train a detector that combines these features to yield a probability map which, after non-maximum suppression, yields a set of medial points. The method is computationally intensive yet parallelizable, and the medial points still need to be parsed and grouped into parts. But the method shows promise in recovering an approximation to a medial axis transform of an image.

The \emph{contour-based} approach is a less holistic approach that addresses the combinatorial challenge of grouping extracted contours. Examples include Brady \& Asada (1984) \cite{Brady84Smoothed}, Connell \& Brady (1987) \cite{Connell87Generating}, Ponce (1990) \cite{Ponce90Characterizing}, Cham \& Cipolla (1995, 1996) \cite{Cham95Symmetry,Cham96Geometric}, Saint-Marc \etal. (1993) \cite{Saintmarc93Bspline}, Liu \etal. (1998) \cite{Liu98Segmenting}, Yl{\"a}-J{\"a}{\"a}ski \& Ade (1996) \cite{Ylajaaski96Grouping}, Stahl \& Wang (2008) \cite{Stahl08Globally}, and Fidler \etal. (2014) \cite{Fidler14Learning}.  Since these methods are contour-based, they have to deal with the issue of computational complexity of contour grouping, particularly when cluttered scenes contain many extraneous edges. Some require smooth contours or initialization, while others were designed to detect symmetric objects and cannot detect and group the symmetric parts that make up an asymmetric object.  A more recent line of methods extract interest point features, such as SIFT \cite{Lowe04Distinctive}, and group them across an unknown symmetry axis \cite{Loy06Detecting, Lee12Curved}.  While these methods exploit distinctive pairwise correspondences among local features, they critically depend on reliable feature extraction.

A recent approach by Narayanan and Kimia \cite{Narayanan12Bottomup} proposes an elegant framework for grouping medial fragments into meaningful groups.  Rather than assuming a figure-ground segmentation, the approach computes a shock graph over the entire image of a cluttered scene, and then applies a sequence of medial transforms to the medial fragments, maintaining a large space of grouping hypotheses.  While the method compares favorably to figure-ground segmentation and fragment generation approaches, the high computational complexity of the approach restricts it to images with no more than 20 contours. 

Our approach, represented in the literature by \cite{Lee13Detecting,Levinshtein09Multiscale,Levinshtein13Multiscale}, is qualitatively different from both filter-based and contour-based approaches, offering a \emph{region-based} approach which perceptually groups together compact regions (segmented at multiple scales using superpixels) representing deformable maximal discs into symmetric parts.  We note that while \cite{Levinshtein09Multiscale} has an additional step that groups symmetric parts into full objects, the scope of our framework is limited to detecting symmetric parts.  In doing so, we avoid the low precision that often plagues the filter-based approaches, along with the high complexity that often plagues the contour-based approaches.

\section{Representing symmetric parts}

Our approach rests on the combination of medial symmetry and superpixel grouping \cite{Lee13Detecting, Levinshtein09Multiscale,Levinshtein13Multiscale}, and in this section we formally connect the two ideas together.  We proceed with the medial axis transform (MAT) \cite{Blum67Transformation} of an object's shape, as illustrated with the runner in Figure \ref{fig:\PartRep}.  The set of maximally inscribed discs plays the central role, whose centers (called \emph{medial points}) trace out the skeleton-like \emph{medial axis} of the object.  We can identify the object's parts by decomposing the medial axis into its branch-like linear segments.  We note that each object part is swept out by the sequence of maximally inscribed discs along the corresponding segment of the medial axis.  For details on the relationship between the medial axis and the simpler reflective axis of symmetry, see Siddiqi \etal. (2008) \cite{Siddiqi08Medial}.

The link between discs and superpixels is established by recently developed approaches that oversegment an image into \emph{superpixels} of compact and uniform scale.  In order to view superpixels as discs, we note that just as superpixels are attracted to parts' boundaries, we imagine removing the circular constraint on discs and allowing them to deform to the boundary, resulting in ``deformable discs''.  We will henceforth use the terms ``superpixel'' and ``disc'' interchangeably.  The disc's shape deforms to the boundary provided that it remains compact (not too long and thin), resulting in a subregion that aligns well with the part's boundary on either side, when such a boundary exists.  In contrast with the maximal disc, which is only bitangent to the boundary, as shown in Figure~\ref{fig:\PartRep}, the number of discs required to compose a part's shape is far less than the number required using maximal discs.

\begin{figure}[t]
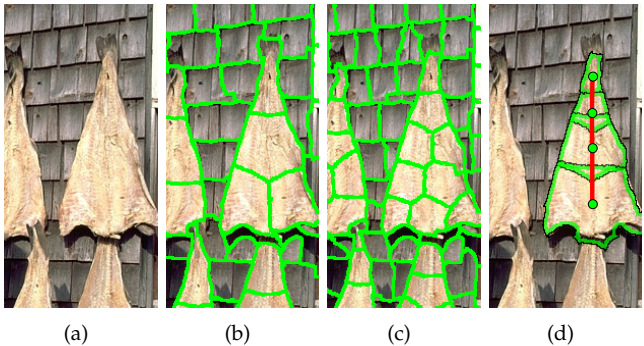

\newcommand{\GetIt}[1]{\includegraphics[height=4cm,clip,trim=0 0 1.5mm 1.5mm]{includegraphics/\Multiscale-#1.pdf}}
\centering
\subfigure[]{\GetIt{00}}
\subfigure[]{\GetIt{03}}
\subfigure[]{\GetIt{04}}
\subfigure[]{\GetIt{05}}
\caption{To compose a part's shape from superpixels in a given input image (a), we compute superpixels at multiple scales (b-c), and combine superpixels from different scales (d).}
\label{fig:\Multiscale}
\end{figure}

In an input image domain of cluttered scenes, the vast majority of superpixels will \emph{not} correspond to true discs of an object's parts, and thus it is suitable to treat superpixels as a set of \emph{candidate} discs.  Furthermore, a superpixel that is too fine or too coarse for a given symmetric part fails to relate its opposing boundaries together into a true disc, and a tapered part may be composed of discs of different sizes, as shown in Figure \ref{fig:\Multiscale}.  Since we have no prior knowledge of a part's scale, and an input image may contain object parts of different scales, we compute superpixels at different scales, and take their union as a set of candidate discs.

Our goal is to perceptually group discs that belong to the same part.  To facilitate grouping decisions, we will define a pairwise affinity function to capture non-accidental relations between discs.  Since the vast majority of superpixels will not correspond to true discs, however, we must manage the complexity of the search space.  By restricting affinities to \emph{adjacent} discs, we exploit one of the most basic grouping cues, namely, \emph{proximity}, which dictates that nearby discs are more likely to belong to the same medial part.  We enlist the help of more sophisticated cues, however, to separate those pairs of discs that belong to the same part from those that do not.  Viewing superpixels as discs allows us to directly exploit the structure of medial symmetry to define the affinity.  In Section \ref{sec:affinity}, we motivate and define the affinity function from perceptual grouping principles to set up a weighted graph $\mathcal{G}$ of disc candidates.  In Section \ref{sec:grouping}, we discuss alternative graph-based algorithms for grouping discs into medial parts.  Section \ref{sec:results} presents qualitative and quantitative experiments, while Section \ref{sec:conclusion} draws some conclusions about the framework.

\section{Disc affinity}
\label{sec:affinity}

Since bottom-up grouping is category-agnostic, a supporting disc affinity must accommodate variations across objects of all types.  The affinity $A(d_i,d_j)$ between discs $d_i$ and $d_j$ must be robust against variability not only within object categories, but also variability between object categories.  For a discriminatively trained affinity, it is helpful to extract features that reduce the variability for the classifier.  In the following sections we define both shape and appearance features on the region scope defined by $d_i$ and $d_j$.

\subsection{Shape features}

\newcommand{\WarpedShape}{WarpedShape}

\begin{figure*}[t]
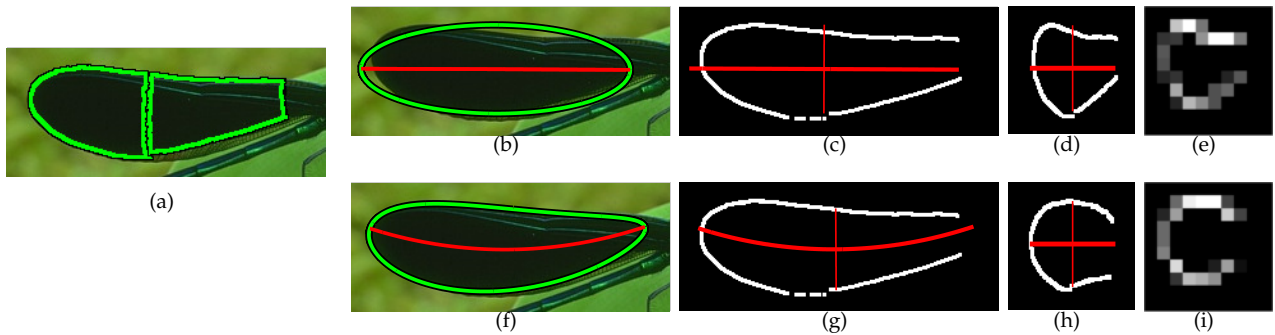

\newcommand{\GetIt}[1]{\includegraphics[height=1.7cm]{includegraphics/\WarpedShape-#1}}
\newcommand{\GetItA}[1]{\includegraphics[height=1.7cm,clip,trim=0 0 1mm 1mm]{includegraphics/\WarpedShape-#1}}
\newcommand{\GetItB}[1]{\includegraphics[height=1.7cm,clip,trim=0 0 2mm 2mm]{includegraphics/\WarpedShape-#1}}
\centering
\subfigure[]{\GetItA{00.pdf}}
\begin{tabular}{c}
\subfigure[]{\GetItA{01.pdf}}
\subfigure[]{\GetItA{03.pdf}}
\subfigure[]{\GetItB{05.pdf}}
\subfigure[]{\GetIt{07.png}}\\
\subfigure[]{\GetItA{02.pdf}}
\subfigure[]{\GetItA{04.pdf}}
\subfigure[]{\GetItB{06.pdf}}
\subfigure[]{\GetIt{08.png}}
\end{tabular}
\caption{Improving invariance with a deformable ellipse: given two adjacent candidate discs, the first step is to fit the ellipse parameters to the region defined by their corresponding superpixels in (a).  The top row shows invariance achieved with a standard ellipse.  The ellipse's fit is visualized with the major axis in (b), the region's boundary edgels before (c) and after (d) warping out the unwanted variations, and the resulting spatial histogram of gradient pixels (e).  See text for details.  The bottom row shows the corresponding steps (f-i) obtained by the deformable ellipse.  Comparing the results, a visually more symmetric feature is obtained by the deformable ellipse, which fits tightly around the region's boundary as compared with the standard ellipse.}
\label{fig:\WarpedShape}
\end{figure*}

The local shape of discs is captured by a spatial histogram of gradient pixels, as illustrated in Figure~\ref{fig:\WarpedShape}.  By encoding the distribution of the boundary edgels of the region defined by the union of the two discs, we capture mid-level shape while avoiding features specific to the given exemplar.  This representation offers us a degree of robustness that is helpful for training the classifier, however it is not perfect---it remains sensitive to variations like scale and orientation, to name a few, and can thus allow the classifier to overfit to training examples.

We turn to medial symmetry to capture these unwanted variations, as the first step in making the feature invariant to such changes.  Specifically, we locally model the shape by fitting the parameters of a symmetric shape to the region.  We refer to a vector $\tu{w}$ of warping parameters that subsequently define a warping function $W:\mathbbm{R}^2\rightarrow\mathbbm{R}^2$ that is used to remove the variations from the space, in effect normalizing the local coordinate system.  Figure~\ref{fig:\WarpedShape} visualizes the parameters $\tu{w}$ of a deformable ellipse fit to a local region, the medial axis before and after the local curvature was ``warped out'' from the coordinate system, and the spatial histogram computed on the normalized coordinate system.

Before describing the spatial histogram in detail, we discuss a class of ellipse-based models for modeling the local medial symmetry.  Ellipses represent ideal shapes of an object's parts, and in particular are shapes that are symmetric about their major axes.  A standard ellipse is parameterized by $\tu{w}_e = (\tu{p},\theta,\tu{a})$, where $\tu{p}$ denotes its position, $\theta$ its orientation, and $\tu{a}=(a_x,a_y)$ the lengths of its major and minor axes.  The parameter vector $\tu{w}_e$ is analytically fit to the local region and is used to define the corresponding warping function $W_e(\tu{w}_e)$.

Historically, we first obtained the warping parameters with an ellipse \cite{Levinshtein09Multiscale}.  While the advantages of using the ellipse lie in its simplicity and ease of fitting, shortcomings were identified in its tendency to provide too coarse a fit to the boundary to yield an accurate enough warping function.  Accordingly in \cite{Lee13Detecting}, we added deformation parameters to obtain a better overall fit across all examples.  Despite a higher computational cost of fitting, the deformable model was shown to yield quantitative improvements.

Specifically, we obtain invariance to bending and tapering deformations by augmenting the ellipse parameters as follows: $\tu{w}_d = (\tu{p},\theta,\tu{a},b,t)$ with the bending radius $b$ along the major axis and tapering slope $t$ along the major axis.  The parameter vector $\tu{w}_d$ is fit by initializing as a standard ellipse and iteratively fitting it to the local region's boundary with a non-linear least-squares algorithm.  The fitted parameters are then used to define the warping function $W_d(\tu{w}_d)$ corresponding to the deformable ellipse.

Only once the warping function $W(\tu{w})$ is fit to the local region and applied to normalize the local coordinate system do we compute the spatial histogram feature.  We place a $10 \times 10$ grid on the warped region, and focusing on the model fit to the union of the two discs, we scale the grid to cover the area $[-1.5a_x,1.5a_x] \times [-1.5a_y,1.5a_y]$.  Using the grid, we compute a 2D histogram on the normalized boundary coordinates weighted by the edge strength of each boundary pixel.  Figure~\ref{fig:\WarpedShape} illustrates the shape feature computed for the disc pair.  We train a SVM classifier on this 100-dimensional feature using our manually labeled superpixel pairs, labeled as belonging to the same part or not. The margin from the classifier is fed into a logistic regressor in order to obtain the shape affinity $A_{shape}(d_i,d_j)$ in the range [0,1].

\subsection{Appearance features}

Aside from medial symmetry, we include appearance similarity as an additional grouping cue.  While object parts may vary widely in color and texture, regions of similar appearance tend to belong to the same part.  We extract an appearance feature on the discs $d_i,d_j$ that encodes their dissimilarity in color and texture.  Specifically, we compute the absolute difference in mean RGB color, absolute difference in mean HSV color, RGB and HSV color variances in both discs, and histogram distance in HSV space, yielding a 27-dimensional appearance feature.  To improve classification, we compute quadratic kernel features, resulting in a 406-dimensional appearance feature.  We train a logistic regressor with L1-regularization to prevent overfitting on a relatively small dataset, while emphasizing the weights of more important features.  This yields an appearance affinity function between two discs $A_{app}(d_i,d_j)$.  Training the appearance affinity is easier than training the shape affinity.  For positive examples, we choose pairs of adjacent superpixels that are contained inside a figure in the figure-ground segmentation, whereas for negative examples, we choose pairs of adjacent superpixels that span figure-ground boundaries.

We combine the shape and appearance affinities using a logistic regressor to obtain the final pairwise affinity $A(d_i,d_j)$.  Both the shape and the appearance affinities, as well as the final affinity $A(d_i,d_j)$, were trained with a regularization parameter of 0.5 on the L1-norm of the logistic coefficients.

\section{Grouping discs}
\label{sec:grouping}

\newcommand{\GraphGrouping}{GraphGrouping}
\newcommand{\SeqOpt}{SeqOpt}

Given a graph $\mathcal{G}$ of discs weighted by affinities, the final step is to group discs that belong to the same symmetric part.  If two adjacent discs correspond to medial points belonging to the same medial axis, they can be combined to extend the symmetry. This is the basis for defining the pairwise affinities in $\mathcal{G}$, and it is how we exploit our medial representation of symmetric parts for grouping.  Specifically, the affinity between two adjacent discs reflects the degree to which it is believed that they not only non-accidentally relate the two opposing boundaries together, but that they are centered along the same medial axis.  In this section, we adapt and discuss two alternative graph-based algorithms, namely, the agglomerative clustering algorithm of Felzenszwalb \& Huttenlocher (2004) \cite{Felzenszwalb04Efficient}, and the sequence-finding algorithm in the salient curve detection method of Felzenszwalb \& McAllester (2006) \cite{Felzenszwalb06Mincover}.
\begin{figure}[t]
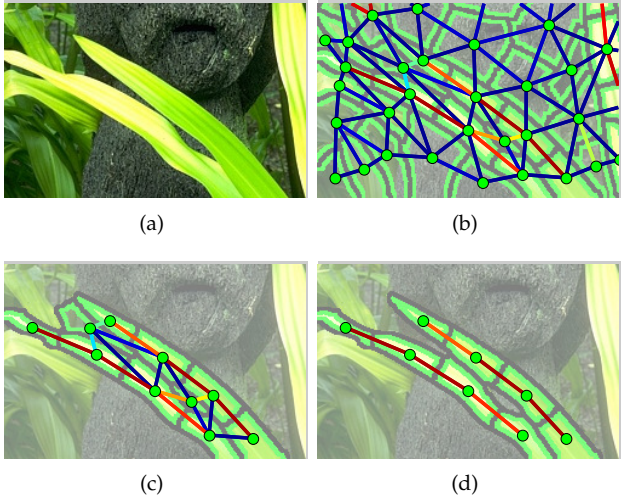

\newcommand{\GetIt}[1]{\includegraphics[height=2.6cm]{includegraphics/\GraphGrouping-#1.pdf}}
\centering
\subfigure[]{\GetIt{00}}
\subfigure[]{\GetIt{01}}
\subfigure[]{\GetIt{02}}
\subfigure[]{\GetIt{03}}

\caption{In our approach, the (a) input image of foreground leaves is oversegmented into superpixels, and a (b) weighted graph $\mathcal{G}$ is built that captures the pairwise affinities that are computed among the superpixels.  A graph-based grouping algorithm takes as input the graph $\mathcal{G}$, which may contain false positive affinities between the leaves, as shown in (b).  In this figure, we illustrate the relative advantage of (d) sequence optimization over (c) agglomerative clustering.  In (c), merging the vertices in $\mathcal{G}$ results in a cluster that undersegments the leaves, combining them into a single symmetric part that violates the assumption that a part is composed of a linear sequence of discs.  In (d), the branching constraint is built into the sequence-finding algorithm which prevents symmetric parts from having tree-structured discs, and correctly segments the leaves into two distinct parts.}
\label{fig:\GraphGrouping}
\end{figure}

\subsection{Agglomerative clustering}

Our first grouping approach is based on agglomerative clustering \cite{Felzenszwalb04Efficient}.  The algorithm takes as input the weighted graph $\mathcal{G}$ and merges edges in increasing order of weights.  Each merge represents a grouping of discs, and the connected components that result correspond to symmetric parts.  Grouping is performed efficiently in $O(e \log e)$ time, where $e$ is the number of edges in $\mathcal{G}$.  We refer the reader to \cite{Levinshtein09Multiscale} for details on the algorithm's adaptation to the setting of grouping discs.

The greedy approach, while fast, is unfortunately underconstrained in allowing merges to occur between branch-structured clusters, resulting in tree-like clusters as illustrated in Figure~\ref{fig:\GraphGrouping}.  These types of clusters can occur as frequently as spuriously high affinity values (false positives) occur, thus motivating the need to constrain the growth of clusters within medial branches.

\subsection{Sequence optimization by dynamic programming}

\begin{figure}[t]
\centering
\includegraphics[height=1.8cm]{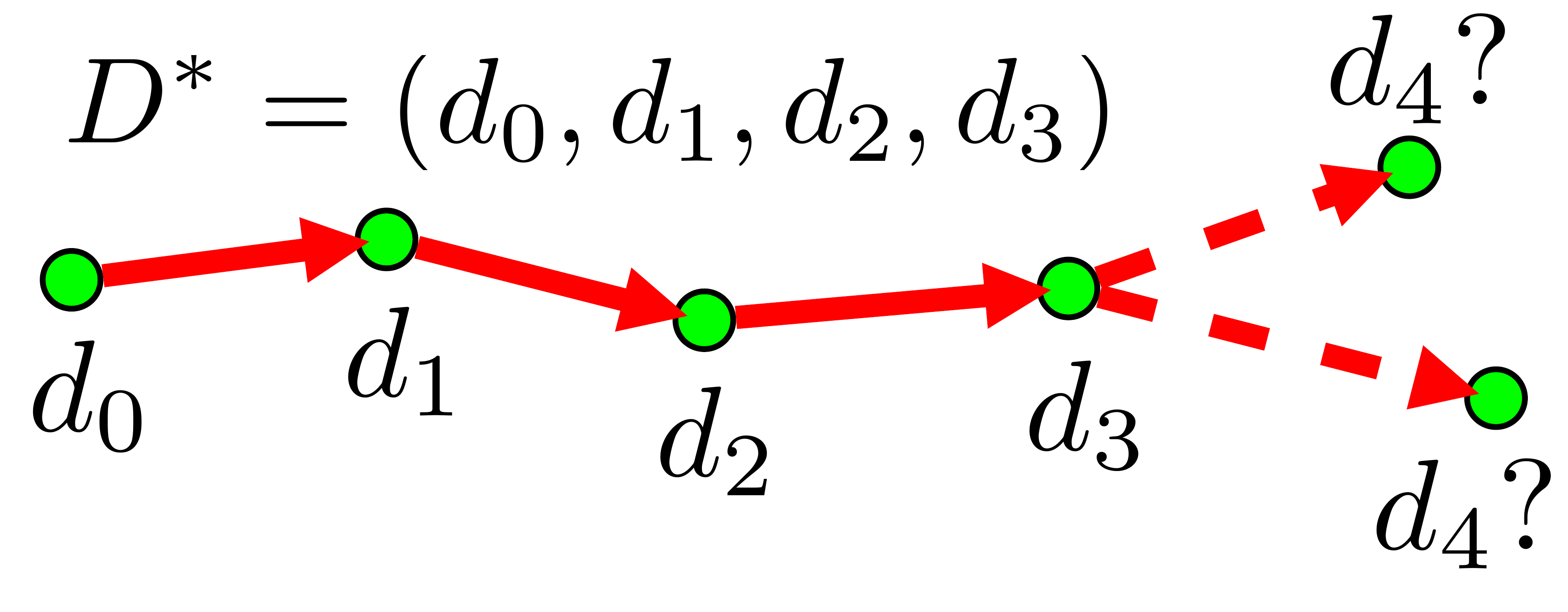}
\caption{Grouping by dynamic programming: The iterative step of the algorithm grows sequences by extracting a sequence $D^*$ from the priority queue, and returning longer sequences to the queue obtained by extending the end of $D^*$ with adjacent discs.  See text for details.
}
\label{fig:\SeqOpt}
\end{figure}

Our second approach is dynamic programming used in \cite{Lee13Detecting}, which observes that each symmetric part is swept out by an \emph{ordered sequence} of discs.  Discs along the same medial axis are thus not only combined in pairs, but can be traced out linearly.  This allows us to reformulate the problem of superpixel grouping as finding sequences of discs in a weighted graph $\mathcal{G}$ that belong to the same symmetric part.  We thus obtain a grouping approach in which the desired branching constraint is inherent in the problem formulation.  As illustrated in Figure~\ref{fig:\GraphGrouping}, the algorithm applied to the same graph prevents the resulting clusters from violating the branching constraint.

Before describing the steps of the dynamic programming algorithm, we note that it solves a discrete optimization problem, and thus represents a principled reformulation of our grouping problem.  This includes defining an objective function that captures the goal of the problem, which is missing from the first approach, and making use of dynamic programming that efficiently solves for a global optimum.  We specifically borrow from the optimization framework used for salient curve detection in Felzenszwalb \& McAllester \cite{Felzenszwalb06Mincover} and adapt it for symmetric part detection.

The application of \cite{Felzenszwalb06Mincover} to our setting is best explained via their method for curve detection.  The method takes as input a graph with weights defined on edges, and ``transition weights'' defined on pairs of adjacent edges.  A salient curve is modeled as a valid sequence of edges, and a regularized cost function is defined on valid sequences that includes a normalized sum of the weights along the given sequence.  Salient curves are found by globally optimizing the cost function using a dynamic programming algorithm.

In our setting, the graph $\mathcal{G}$ supplies weights between adjacent discs, and we define a valid sequence of discs (of variable length) by $D=(d_0,d_1,\ldots,d_n)$, which represents a symmetric part.  The criteria that we want to optimize---good symmetry along the medial axis and a maximally long axis---is provided by the affinity graph $\mathcal{G}$.  The regularized cost function, $\text{cost}(D)$, is defined correspondingly, and favors good internal affinity with a normalized sum over the affinities along the given sequence, and encourages longer sequences with a regularization term.  Affinities defined over longer subsequences corresponding to the transition weights have a smoothing effect on the preferred sequences.  Details on the cost function including its mathematical form can be found in \cite{Lee13Detecting}.

We now summarize the dynamic programming steps for globally minimizing $\text{cost}(D)$.  The core step is illustrated in Figure~\ref{fig:\SeqOpt}, and details can be found in \cite{Felzenszwalb06Mincover}.  The algorithm initializes a priority queue $Q$ of candidate sequences with all possible sequences of unit length, then pursues a best-first search strategy of iteratively extending the cheapest candidate sequences.  Each edge $(d_{i-1},d_i)$ is directed such that a sequence of edges terminating at $d_i$ can be extended with an edge starting at $d_i$.  At each iteration, as shown in Figure~\ref{fig:\SeqOpt}, the most promising sequence $D^*$ is removed from $Q$, and new candidate sequences are proposed by extending the end of $D^*$ with adjacent discs.  If an extended sequence ending at an edge improves the cost of an existing sequence ending at the same edge, it is added back into $Q$.  To find multiple sequences from the graph corresponding to different symmetric parts, we iteratively remove sequences that are already found and re-minimize the cost, until a maximum cost is reached.

\section{Results}
\label{sec:results}

\newcommand{\MultipleDetections}{MultipleDetections-dark}
\newcommand{\QualResults}{QualResults-color-dark}

\begin{figure}[t]
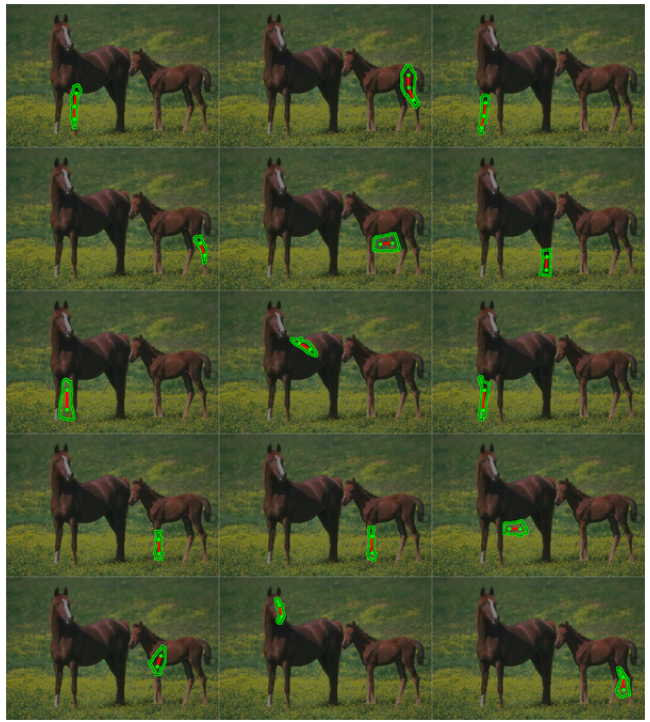
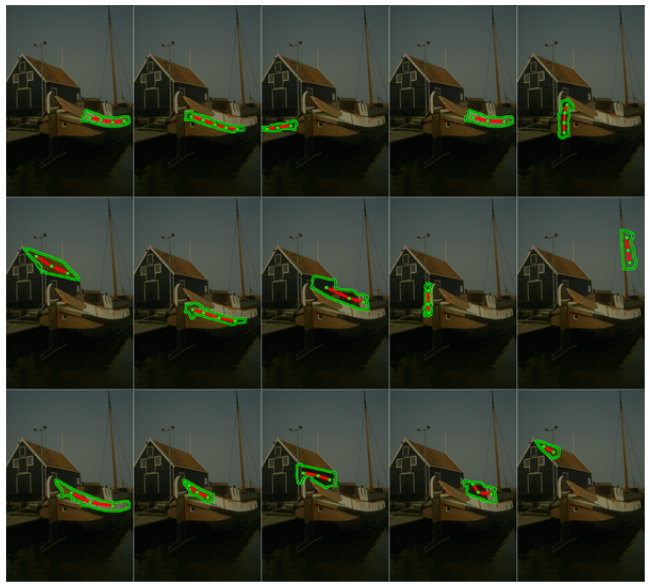

\newcommand{\GetIt}[1]{\includegraphics[width=\linewidth]{includegraphics/\MultipleDetections-#1.png}}
\centering
\subfigure[]{\GetIt{00}}
\subfigure[]{\GetIt{01}}
\caption{Multiple symmetric parts: for each image (a), (b) below we show the top 15 masks detected as symmetric parts.  Each mask is detected as a sequence of discs, whose centers are plotted in green and connected by a sequence of red line segments that represents the medial axis.}
\label{fig:\MultipleDetections}
\end{figure}

We present an evaluation of our approach, first qualitatively in Section \ref{sec:qualitative}, then quantitatively in Section \ref{sec:quantitative}.  Our qualitative results are drawn from sample input images and illustrate particular strengths and weaknesses of our approach.  In our quantitative evaluation, we use performance metrics on two different datasets to gauge the contributions of different components in our approach.  Figure \ref{fig:\MultipleDetections} visualizes detected masks returned by our method, specifically showing the top 15 detected parts on sample input images.  Parts are ranked by the optimization objective function.  On each part's mask, we indicate the associated disc centers and the medial axis via connecting line segments.  All results reported are generated with superpixels computed using normalized cuts \cite{Shi00Normalized}, at multiple scales corresponding to 25, 50, 100, and 200 superpixels per image.

Our evaluation employs two image datasets of cluttered scenes.  The first dataset is a subset of 81 images from the Weizmann Horse Database (WHD) \cite{Borenstein02Classspecific}, in which each image contains one or more horses.  Aside from color variation, the dataset exhibits variations in scale, position, articulation of horse joints.  The second dataset was created by Lee \etal. \cite{Lee13Detecting} from the Berkeley Segmentation Database (BSD) \cite{Martin01Database}.  This set is denoted as BSD-Parts and contains 36 BSD images which are annotated with ground-truth masks corresponding to the symmetric parts of prominent objects (\eg, duck, horse, deer, snake, boat, dome, amphitheater).  This contains a variety of natural and artificial objects and offers a balancing counterpart to the horse dataset.

Both WHD and BSD-Parts are annotated with ground-truth masks corresponding to object parts in the image.  The learning component of our approach requires ground-truth masks as input, for which we have held a subset of training images away from testing.  Specifically, we trained our classifier on 20 WHD images and used for evaluation the remaining 61 WHD images and all 36 BSD-Parts images.  This methodology supports a key point of our approach, which is that of \emph{mid-level transfer}: by increasing feature invariance against image variability, we help prevent the classifier from overfitting to the objects on which it is trained.  By training our model on horse images and applying it on other types of objects, we thus demonstrate the ability of our model to transfer symmetric part detection from one object class to another.

\subsection{Qualitative results}
\label{sec:qualitative}

\begin{figure}[t]
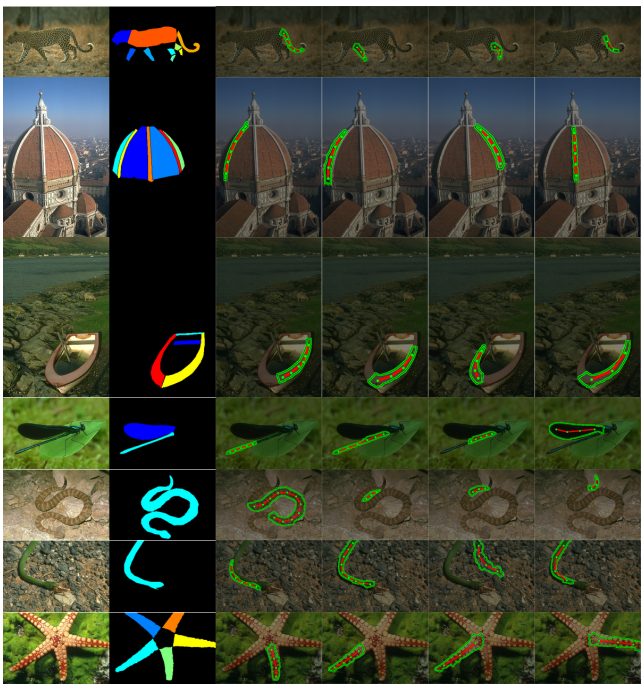

\newcommand{\pheight}{1.605cm}
\newcommand{\lheight}{3.60cm}
\newcommand{\GetIt}[2]{\includegraphics[width=\linewidth]{includegraphics/\QualResults-#1.png}}
\centering
\GetIt{11}{\pheight} 
\GetIt{02}{\lheight}
\GetIt{31}{\lheight} 
\GetIt{04}{\pheight}
\GetIt{30}{\pheight} 
\GetIt{34}{\pheight} 
\GetIt{01}{\pheight} 
\caption{Example detections on a sample of images from BSD-Parts.  Columns left to right: input image, ground-truth masks, top 4 detection masks.  Note that many images have more ground-truth masks than detections that can be shown here.}
\label{fig:\QualResults}
\end{figure}

Figure~\ref{fig:\QualResults} presents our results on a sample of input images.  For each image, the set of ground-truth masks are shown, followed by the top several detection masks.  (Detection masks are indicated with the associated sequence of discs.)  For clarity, individual detections are shown in separate images.  The tiger image demonstrates successful detection of its parts, which vary in curvature and taper.  In the next example, vertical segments of the Florentine dome are detected by the same method.  The next example shows recovered parts of the boat.  When suitably pruned, a configuration of parts hypothesized from a cluttered image can provide an index into a bank of part-based shape models.

In the image of the fly, noise along the abdomen was captured by the affinity function at finer superpixel scales, resulting in multiple overlapping oversegmentations.  The leaf was not detected, however, due to its symmetry being occluded.  In the first snake image, low contrast along its tail yielded imperfect superpixels that could not support correct segmentation, however the invariance to bending is impressive.  The second snake is accompanied with a second thin detection along its shadow.  We conclude with the starfish whose complex texture was not a difficult challenge for our method.  We have demonstrated that symmetry is a powerful shape regularity that is ubiquitous across different objects.

\subsection{Quantitative results}
\label{sec:quantitative}

\begin{figure}[t]
\subfigure[BSD-Parts]{
\includegraphics[width=\linewidth]{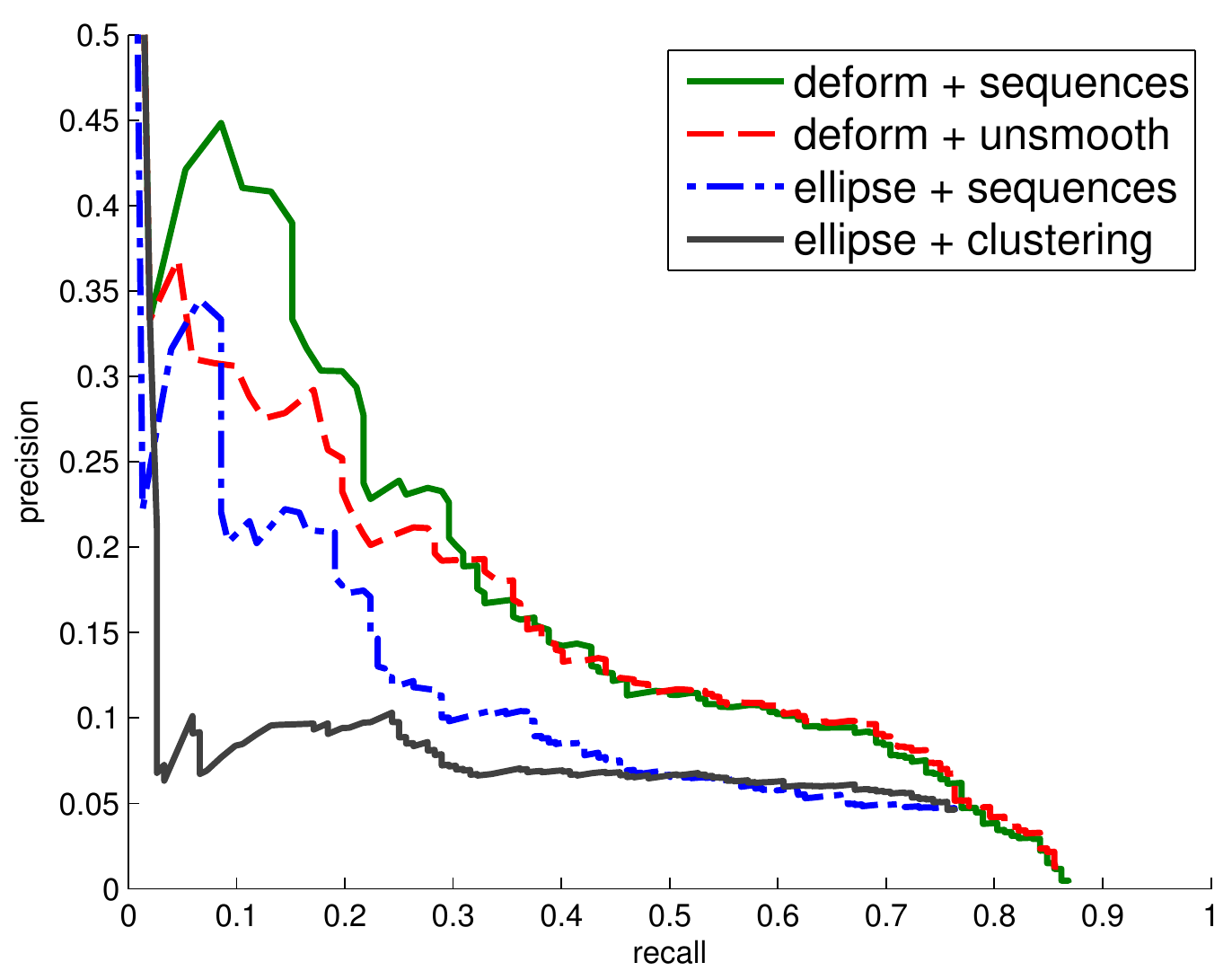}
\label{fig:eval-pr-one-BSDS}
}
\subfigure[WHD]{
\includegraphics[width=\linewidth]{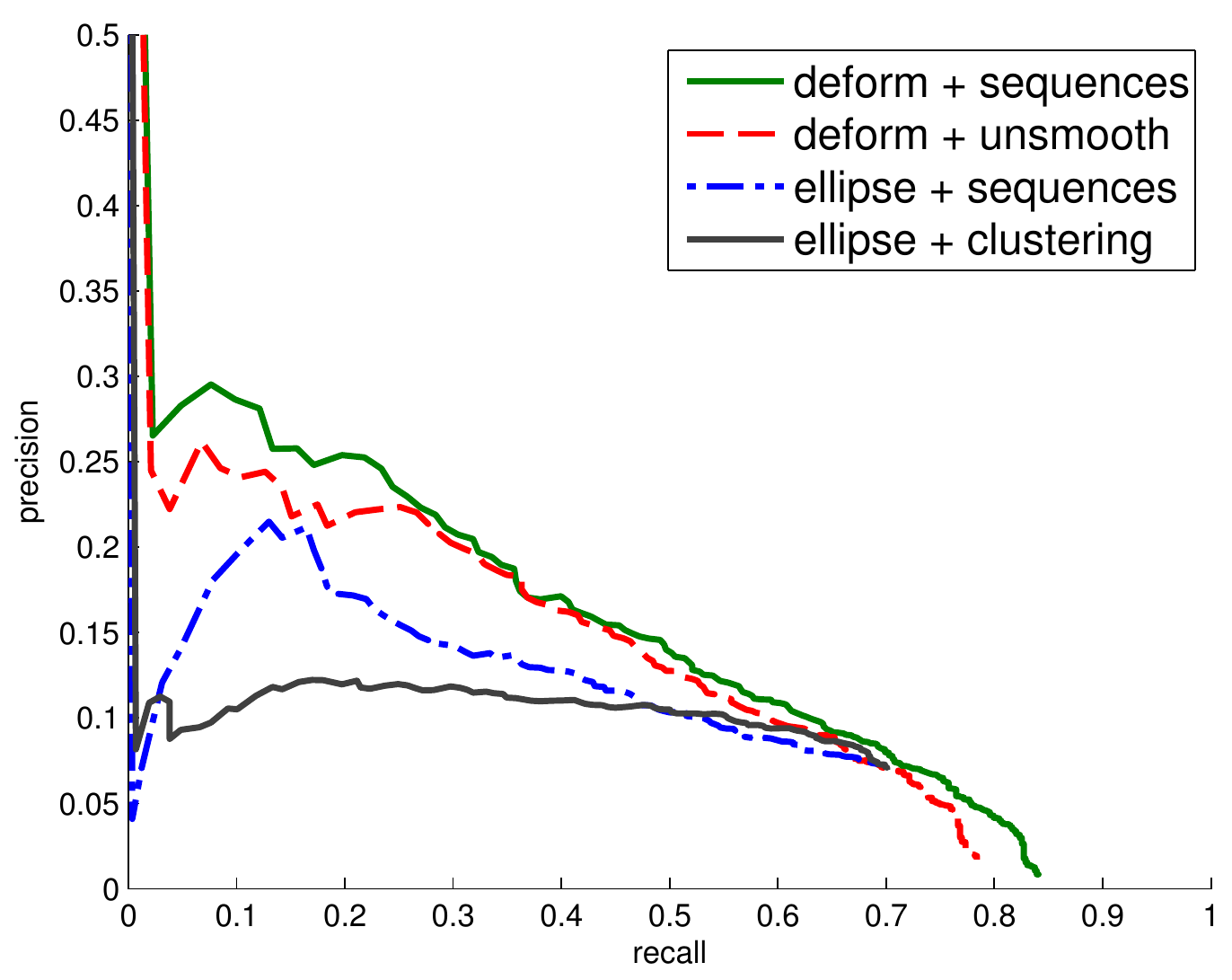}
\label{fig:eval-pr-one-WHD}
}
\caption{Performance curves for corresponding to different settings of the components of our approach on (a) BSD-Parts and (b) WHD.  See text for details.}
\label{fig:eval-pr}
\end{figure}

In the quantitative part of our evaluation, we use standard dataset metrics to evaluate the components of our approach.  Specifically, we demonstrate the improvement contributed by formulating grouping as sequence optimization, and by using invariant features to train the classifier.  Results are computed on the subset of WHD held out from training, and on BSD-Parts.  To evaluate the quality of our detected symmetric parts, we compare them in the form of detection masks to the ground-truth masks using the standard intersection-over-union metric (IoU).  A detection mask $m_{det}$ is counted as a hit if its overlap with the ground-truth mask $m_{gt}$ is greater than 0.4, where overlap is measured by IoU $|m_{det} \cap m_{gt}| / |m_{det} \cup m_{gt}|$.  We obtain a precision-recall curve by varying the threshold over the cost (weight) of detected parts.

Figure~\ref{fig:eval-pr} presents the performance curves corresponding to 4 different settings under our framework, evaluated on both WHD and BSD-Parts: 1) \emph{ellipse+clustering} combines the ellipse-warped affinity with agglomerative clustering and corresponds to \cite{Levinshtein09Multiscale}.  We note that low precision is partly due to the lack of annotations on many background objects in both datasets; 2) \emph{ellipse+sequences} combines the ellipse-warped affinity with sequence optimization; 3) \emph{deform+sequences} combines deformable warping with sequence optimization, and corresponds to \cite{Lee13Detecting}; and 4) \emph{deform+unsmooth} sets the triplewise weights in $\text{cost}(D)$ uniformly to zero rather than using the affinity as done in the previous setting.  A corresponding drop in performance shows that smoothness is an important feature of symmetric parts.  In summary, experimental results confirm that both the added deformations and sequence optimization are individually effective at improving the accuracy of our approach.

\section{Conclusion}
\label{sec:conclusion}

Symmetry figured prominently in early object recognition systems, but the potential of this powerful cue is largely overlooked in contemporary computer vision.  In this article, we have reviewed a framework that attempts to reintroduce medial symmetry into the current research landscape.  
The key concept behind the framework is remodeling the discs of the MAT as compact superpixels, learning a pairwise affinity function between discs with a symmetry-invariant transform, and formulating a discrete optimization problem to find the best sequences of discs.  We have summarized quantitative results that encourage further exploration of using symmetry for object recognition.

We have reviewed ways in which we overcame the early limitations of our approach, such as using additional deformation parameters to improve warping accuracy, and reformulating grouping as a discrete optimization problem to improve results.  There are also current limitations to be addressed in future work.  To briefly mention two, we first note that the success of using Gestalt grouping cues such as symmetry depends on effectively combining multiple cues together.  To improve the robustness of our system, we are thus exploring how to incorporate additional mid-level cues such as contour closure.  This will help our system more accurately resolve cases where different features provide conflicting cues, and thus improve the overall performance.  Secondly, our scope is bottom-up detection and thus is agnostic of object categories.  However, in a detection or verification task, top-down cues may be available.  We are thus investigating ways of integrating top-down cues into our framework.  

In conclusion, we have reviewed an approach for reintroducing the MAT back into contemporary computer vision, by leveraging the formulation of maximal discs as compact superpixels to derive symmetry-based affinity function and grouping algorithms.  Quantitative results encourage further development of the framework to recover medial-based parts from cluttered scenes. Finally, as initial explored in \cite{Levinshtein13Multiscale}, detected parts must be non-accidentally grouped before they yield the distinctiveness required for object recognition.

\ifCLASSOPTIONcompsoc
  \section*{Acknowledgments}
\else
  \section*{Acknowledgment}
\fi

We thank Allan Jepson for valuable discussions on dynamic programming algorithms.

\ifCLASSOPTIONcaptionsoff
  \newpage
\fi



\bibliographystyle{IEEEtran}
\bibliography{SymmetricParts_arxiv}

\begin{thebibliography}{10}
\providecommand{\url}[1]{#1}
\csname url@samestyle\endcsname
\providecommand{\newblock}{\relax}
\providecommand{\bibinfo}[2]{#2}
\providecommand{\BIBentrySTDinterwordspacing}{\spaceskip=0pt\relax}
\providecommand{\BIBentryALTinterwordstretchfactor}{4}
\providecommand{\BIBentryALTinterwordspacing}{\spaceskip=\fontdimen2\font plus
\BIBentryALTinterwordstretchfactor\fontdimen3\font minus
  \fontdimen4\font\relax}
\providecommand{\BIBforeignlanguage}[2]{{%
\expandafter\ifx\csname l@#1\endcsname\relax
\typeout{** WARNING: IEEEtran.bst: No hyphenation pattern has been}%
\typeout{** loaded for the language `#1'. Using the pattern for}%
\typeout{** the default language instead.}%
\else
\language=\csname l@#1\endcsname
\fi
#2}}
\providecommand{\BIBdecl}{\relax}
\BIBdecl

\bibitem{Lee13Detecting}
\BIBentryALTinterwordspacing
T.~Lee, S.~Fidler, and S.~Dickinson, ``Detecting curved symmetric parts using a
  deformable disc model,'' \emph{ICCV}, 2013.
\BIBentrySTDinterwordspacing

\bibitem{Levinshtein09Multiscale}
A.~Levinshtein, S.~Dickinson, and C.~Sminchisescu, ``Multiscale symmetric part
  detection and grouping,'' \emph{ICCV}, 2009.

\bibitem{Levinshtein13Multiscale}
A.~Levinshtein, C.~Sminchisescu, and S.~Dickinson, ``Multiscale symmetric part
  detection and grouping,'' \emph{IJCV}, vol. 104, no.~2, pp. 117--134, 2013.

\bibitem{Leyton92Symmetry}
M.~Leyton, ``Symmetry, causality, mind,'' \emph{MIT Press}, 1992.

\bibitem{Wertheimer38Laws}
\BIBentryALTinterwordspacing
M.~Wertheimer, ``Laws of organization in perceptual forms,'' \emph{Source Book
  of Gestalt Psychology}, Jan 1938.
\BIBentrySTDinterwordspacing

\bibitem{Tyler02Human}
\BIBentryALTinterwordspacing
C.~Tyler, ``Human symmetry perception and its computational analysis,''
  \emph{Taylor {\&} Francis}, Jan 2002.
\BIBentrySTDinterwordspacing

\bibitem{Blum67Transformation}
H.~Blum, ``A transformation for extracting new descriptors of shape,''
  \emph{Models for the perception of speech and visual form}, vol.~19, no.~5,
  pp. 362--380, 1967.

\bibitem{Binford71Visual}
T.~Binford, ``Visual perception by computer,'' \emph{ICSC}, 1971.

\bibitem{Pentland86Perceptual}
\BIBentryALTinterwordspacing
A.~Pentland, ``Perceptual organization and the representation of natural
  form,'' \emph{Artificial Intelligence}, vol.~28, no.~3, pp. 293--331, 1986.
\BIBentrySTDinterwordspacing

\bibitem{Biederman85Human}
\BIBentryALTinterwordspacing
I.~Biederman, ``Human image understanding: Recent research and a theory,''
  \emph{CVGIP}, Jan 1985.
\BIBentrySTDinterwordspacing

\bibitem{Siddiqi99Shock}
\BIBentryALTinterwordspacing
K.~Siddiqi, A.~Shokoufandeh, S.~Dickinson, and S.~Zucker, ``Shock graphs and
  shape matching,'' \emph{IJCV}, vol.~35, no.~1, pp. 13--32, 1999.
\BIBentrySTDinterwordspacing

\bibitem{Sebastian04Recognition}
T.~Sebastian, P.~Klein, and B.~Kimia, ``Recognition of shapes by editing their
  shock graphs,'' \emph{PAMI}, vol.~26, no.~5, pp. 550--571, Jul 2004.

\bibitem{Siddiqi08Medial}
\BIBentryALTinterwordspacing
K.~Siddiqi and S.~Pizer, ``Medial representations: mathematics, algorithms and
  applications,'' \emph{Springer}, Jan 2008.
\BIBentrySTDinterwordspacing

\bibitem{Girshick14Rich}
\BIBentryALTinterwordspacing
R.~Girshick, J.~Donahue, T.~Darrell, and J.~Malik, ``Rich feature hierarchies
  for accurate object detection and semantic segmentation,'' \emph{CVPR}, Jan
  2014.
\BIBentrySTDinterwordspacing

\bibitem{Dickinson09Evolution}
\BIBentryALTinterwordspacing
S.~Dickinson, ``The evolution of object categorization and the challenge of
  image abstraction,'' \emph{Object categorization: computer and human vision
  perspectives}, pp. 1--37, Jan 2009.
\BIBentrySTDinterwordspacing

\bibitem{Dickinson13Role}
\BIBentryALTinterwordspacing
S.~Dickinson, A.~Levinshtein, P.~Sala, and C.~Sminchisescu, ``The role of
  mid-level shape priors in perceptual grouping and image abstraction,''
  \emph{Shape Perception in Human and Computer Vision: An Interdisciplinary
  Perspective}, Jan 2013.
\BIBentrySTDinterwordspacing

\bibitem{Levinshtein12Optimal}
A.~Levinshtein, C.~Sminchisescu, and S.~Dickinson, ``Optimal image and video
  closure by superpixel grouping,'' \emph{IJCV}, 2012.

\bibitem{Ren05Cue}
X.~Ren, C.~Fowlkes, and J.~Malik, ``Cue integration for figure/ground
  labeling,'' \emph{NIPS}, 2005.

\bibitem{Lee14Multicue}
\BIBentryALTinterwordspacing
T.~Lee, S.~Fidler, and S.~Dickinson, ``Multi-cue mid-level grouping,''
  \emph{ACCV}, 2014.
\BIBentrySTDinterwordspacing

\bibitem{Pelillo99Matching}
\BIBentryALTinterwordspacing
M.~Pelillo, K.~Siddiqi, and S.~Zucker, ``Matching hierarchical structures using
  association graphs,'' \emph{PAMI}, vol.~21, no.~11, pp. 1105--1120, Jan 1999.
\BIBentrySTDinterwordspacing

\bibitem{Demirci09Skeletal}
\BIBentryALTinterwordspacing
F.~Demirci, A.~Shokoufandeh, and S.~Dickinson, ``Skeletal shape abstraction
  from examples,'' \emph{PAMI}, vol.~31, no.~5, p. 944, 2009.
\BIBentrySTDinterwordspacing

\bibitem{Macrini11Object}
\BIBentryALTinterwordspacing
D.~Macrini, S.~Dickinson, D.~Fleet, and K.~Siddiqi, ``Object categorization
  using bone graphs,'' \emph{CVIU}, Jan 2011.
\BIBentrySTDinterwordspacing

\bibitem{Macrini11Bone}
\BIBentryALTinterwordspacing
------, ``Bone graphs: Medial shape parsing and abstraction,'' \emph{CVIU}, Apr
  2011.
\BIBentrySTDinterwordspacing

\bibitem{Crowley84Representation}
J.~Crowley and A.~Parker, ``A representation for shape based on peaks and
  ridges in the difference of low-pass transform,'' \emph{PAMI}, vol.~6, no.~2,
  pp. 156--170, 1984.

\bibitem{Shokoufandeh99Viewbased}
\BIBentryALTinterwordspacing
A.~Shokoufandeh, I.~Marsic, and S.~Dickinson, ``View-based object recognition
  using saliency maps,'' \emph{IVC}, vol.~17, no. 5-6, pp. 445--460, 1999.
\BIBentrySTDinterwordspacing

\bibitem{Mikolajczyk02Affine}
K.~Mikolajczyk and C.~Schmid, ``An affine invariant interest point detector,''
  \emph{ECCV}, 2002.

\bibitem{Lindeberg03Realtime}
\BIBentryALTinterwordspacing
T.~Lindeberg and L.~Bretzner, ``Real-time scale selection in hybrid multi-scale
  representations,'' \emph{Scale Space Methods in Computer Vision}, Jan 2003.
\BIBentrySTDinterwordspacing

\bibitem{Shokoufandeh06Representation}
A.~Shokoufandeh, L.~Bretzner, D.~Macrini, M.~F. Demirci, C.~J{\"o}nsson, and
  S.~Dickinson, ``The representation and matching of categorical shape,''
  \emph{CVIU}, vol. 103, no.~2, pp. 139--154, 2006.

\bibitem{Tsogkas12Learningbased}
\BIBentryALTinterwordspacing
S.~Tsogkas and I.~Kokkinos, ``Learning-based symmetry detection in natural
  images,'' \emph{ECCV}, Jan 2012.
\BIBentrySTDinterwordspacing

\bibitem{Brady84Smoothed}
M.~Brady and H.~Asada, ``Smoothed local symmetries and their implementation,''
  \emph{IJRR}, vol.~3, no.~3, pp. 36--61, 1984.

\bibitem{Connell87Generating}
\BIBentryALTinterwordspacing
J.~Connell and M.~Brady, ``Generating and generalizing models of visual
  objects,'' \emph{Artificial Intelligence}, Jan 1987.
\BIBentrySTDinterwordspacing

\bibitem{Ponce90Characterizing}
J.~Ponce, ``On characterizing ribbons and finding skewed symmetries,''
  \emph{CVGIP}, vol.~52, no.~3, pp. 328--340, 1990.

\bibitem{Cham95Symmetry}
T.-J. Cham and R.~Cipolla, ``Symmetry detection through local skewed
  symmetries,'' \emph{IVC}, vol.~13, no.~5, pp. 439--450, 1995.

\bibitem{Cham96Geometric}
\BIBentryALTinterwordspacing
T.~Cham and R.~Cipolla, ``Geometric saliency of curve correspondences and
  grouping of symmetric contours,'' \emph{ECCV}, Jan 1996.
\BIBentrySTDinterwordspacing

\bibitem{Saintmarc93Bspline}
P.~Saint-Marc, H.~Rom, and G.~Medioni, ``B-spline contour representation and
  symmetry detection,'' \emph{PAMI}, vol.~15, no.~11, pp. 1191--1197, 1993.

\bibitem{Liu98Segmenting}
\BIBentryALTinterwordspacing
T.~Liu, D.~Geiger, and A.~Yuille, ``Segmenting by seeking the symmetry axis,''
  \emph{ICPR}, Jan 1998.
\BIBentrySTDinterwordspacing

\bibitem{Ylajaaski96Grouping}
A.~Yl{\"a}-J{\"a}{\"a}ski and F.~Ade, ``Grouping symmetrical structures for
  object segmentation and description,'' \emph{CVIU}, vol.~63, no.~3, pp.
  399--417, 1996.

\bibitem{Stahl08Globally}
J.~Stahl and S.~Wang, ``Globally optimal grouping for symmetric closed
  boundaries by combining boundary and region information,'' \emph{PAMI},
  vol.~30, no.~3, pp. 395--411, 2008.

\bibitem{Fidler14Learning}
\BIBentryALTinterwordspacing
S.~Fidler, M.~Boben, and A.~Leonardis, ``Learning a hierarchical compositional
  shape vocabulary for multi-class object representation,''
  \emph{ArXiv:1408.5516}, Jan 2014.
\BIBentrySTDinterwordspacing

\bibitem{Lowe04Distinctive}
D.~Lowe, ``Distinctive image features from scale-invariant keypoints,''
  \emph{IJCV}, vol.~60, no.~2, pp. 91--110, 2004.

\bibitem{Loy06Detecting}
G.~Loy and J.~Eklundh, ``Detecting symmetry and symmetric constellations of
  features,'' \emph{ECCV}, 2006.

\bibitem{Lee12Curved}
S.~Lee and Y.~Liu, ``Curved glide-reflection symmetry detection,'' \emph{PAMI},
  vol.~34, no.~2, pp. 266--278, 2012.

\bibitem{Narayanan12Bottomup}
\BIBentryALTinterwordspacing
M.~Narayanan and B.~Kimia, ``Bottom-up perceptual organization of images into
  object part hypotheses,'' \emph{ECCV}, Jan 2012.
\BIBentrySTDinterwordspacing

\bibitem{Felzenszwalb04Efficient}
\BIBentryALTinterwordspacing
P.~Felzenswalb and D.~Huttenlocher, ``Efficient graph-based image
  segmentation,'' \emph{IJCV}, vol.~59, no.~2, pp. 167--181, 2004.
\BIBentrySTDinterwordspacing

\bibitem{Felzenszwalb06Mincover}
P.~Felzenszwalb and D.~McAllester, ``A min-cover approach for finding salient
  curves,'' \emph{WPOCV}, 2006.

\bibitem{Shi00Normalized}
\BIBentryALTinterwordspacing
J.~Shi and J.~Malik, ``Normalized cuts and image segmentation,'' \emph{PAMI},
  vol.~22, no.~8, pp. 888--905, 2000.
\BIBentrySTDinterwordspacing

\bibitem{Borenstein02Classspecific}
\BIBentryALTinterwordspacing
E.~Borenstein and S.~Ullman, ``Class-specific, top-down segmentation,''
  \emph{ECCV}, Jan 2002.
\BIBentrySTDinterwordspacing

\bibitem{Martin01Database}
\BIBentryALTinterwordspacing
D.~Martin, C.~Fowlkes, D.~Tal, and J.~Malik, ``A database of human segmented
  natural images and its application to evaluating segmentation algorithms and
  measuring ecological statistics,'' \emph{ICCV}, Jan 2001.
\BIBentrySTDinterwordspacing

\end{thebibliography}
\end{document}